\documentclass{article}
\usepackage{ijcai23}

\usepackage{times}
\usepackage{soul}
\usepackage{url}
\usepackage[hidelinks]{hyperref}
\usepackage[utf8]{inputenc}
\usepackage[small]{caption}
\usepackage{graphicx}
\usepackage{amsmath}
\usepackage{amsthm}
\usepackage{booktabs}
\usepackage{algorithm}
\usepackage{algorithmic}
\usepackage[switch]{lineno}

\usepackage{diagbox}
\usepackage{array}
\usepackage{comment}
\usepackage{amssymb}
\usepackage{stfloats}
\usepackage[table]{xcolor}
\usepackage{multirow}
\newcolumntype{P}[1]{>{\centering\arraybackslash}p{#1}}
\newcolumntype{M}[1]{>{\centering\arraybackslash}m{#1}}

\newtheorem{example}{Example}
\newtheorem{theorem}{Theorem}

\title{Gradient-Guided Knowledge Distillation for Object Detectors}

\author{
Qizhen Lan$^1$
\and
Qing Tian$^1$
\affiliations
$^1$Department of Computer Science,\\ Bowling Green State University\\
\emails
\{qlan, qtian\}@bgsu.edu,
}

\begin{document}

\maketitle

\begin{abstract}
     Deep learning models have demonstrated remarkable success in object detection, yet their complexity and computational intensity pose a barrier to deploying them in real-world applications (e.g., self-driving perception). Knowledge Distillation (KD) is an effective way to derive efficient models. However, only a small number of KD methods tackle object detection. Also, most of them focus on mimicking the plain features of the teacher model but rarely consider how the features contribute to the final detection. In this paper, we propose a novel approach for knowledge distillation in object detection, named Gradient-guided Knowledge Distillation (GKD). Our GKD uses gradient information to identify and assign more weights to features that significantly impact the detection loss, allowing the student to learn the most relevant features from the teacher. Furthermore, we present bounding-box-aware multi-grained feature imitation (BMFI) to further improve the KD performance. Experiments on the KITTI and COCO-Traffic datasets demonstrate our method's efficacy in knowledge distillation for object detection. On one-stage and two-stage detectors, our GKD-BMFI leads to an average of 5.1\% and 3.8\% mAP improvement, respectively, beating various state-of-the-art KD methods.
\end{abstract}

\section{Introduction}

Over the past few years, deep learning models have achieved remarkable success in a variety of domains, including computer vision \cite{he2016deep,he2017mask,ronneberger2015u}. Object detection is one of the most critical tasks in computer vision and has seen growing demand in various applications, such as autonomous driving, surveillance, and medical imaging. However, high detection performance often comes at the cost of large and complex neural architectures, which results in slow inference speed on devices without powerful GPUs. To address this problem, various neural network compression techniques have been proposed, such as pruning \cite{frankle2018lottery,tian2021task}, quantization \cite{nagel2019data,li2019fully}, and knowledge distillation \cite{hinton2015distilling,li2017mimicking}. In Knowledge Distillation (KD), a smaller, lightweight student model mimics the behavior of an unwieldy pre-trained teacher model to achieve comparable or even superior results. The information transferred across the models is usually referred to as ``dark knowledge'' due to its blackbox nature. Feature-based KD is one of the most popular KD types, which aims to minimize the difference between the teacher's intermediate feature representations and those of the student.

Most of the existing knowledge distillation methods in computer vision are designed for image classification \cite{hinton2015distilling,li2017mimicking,tian2021task,zagoruyko2016paying}. In the past few years, researchers have started to explore how KD can be effectively applied to object detection. Most state-of-the-art KD methods in object detection use feature-based approaches where the student is trained to mimic the teacher's plain or human-selected features. These methods aim to explore which parts of the teacher's features provide the most informative knowledge for the student to distill. For example, \cite{sun2020distilling} and \cite{wang2019distilling} respectively use the Gaussian Mask and the ``fine-grained'' imitation mask to select a broader distillation area. \cite{guo2021distilling} distills the foreground and background separately. \cite{zhang2021improve,yang2022focal} leverage highly activated features and non-local modules to guide the student and distill the global relation of pixels, respectively.
However, few studies have considered how these features contribute to the final detection outcome. Unlike previous approaches, we propose a novel gradient-guided knowledge distillation (GKD) method that incorporates gradient information to weigh the importance of features. The gradients of the detection loss function with respect to the model's features provide information about the features' contribution to the final detection performance. By using the task gradients to weigh the importance of features during knowledge distillation, we can effectively transfer knowledge that is more relevant to the task at hand and has a greater impact on the model's performance. To the best of our knowledge, this is the first work that utilizes gradients to weight the importance of features for knowledge distillation in object detection tasks.
Moreover, we argue that foreground objects, including their surrounding pixels with abundant contextual information, should receive special attention during KD. Unlike \cite{wang2019distilling} that distills pixels around the foreground object with fixed weights, we use a top-flattened Gaussian mask to assign the highest weight to the pixels within the ground truth bounding boxes and gradually decrease the weight of surrounding pixels as the distance from the center point increases. We also find that feature imitation at multiple granularities help with the KD.

In summary, the main contributions of this paper are as follows:

\begin{itemize}
    \item We introduce a novel gradient-guided knowledge distillation (GKD) method that utilizes gradient information to weigh the importance of features so that the student model can focus on the more valuable knowledge that is relevant to the final detection. As far as we know, this is the first time that gradients are leveraged as a knowledge filter in knowledge distillation for object detectors.
    \item We present bounding-box-aware multi-grained feature inmitation that takes bounding boxes and their contextual information into consideration during KD and performs distillation along different feature dimensions.
    \item Our KD method's efficacy is tested on both one-stage and two-stage detectors with different backbones on the KITTI and COCO Traffic datasets. GKD outperforms state-of-the-art knowledge distillation methods, achieving an average 4.3 and 3.1 mAP improvement for single-stage and two-stage detectors, respectively. Additionally, when combined with the proposed Boundary-aware Multi-Feature Imitation (BMFI) method, GKD-BMFI achieves an average 4.7 and 3.7 mAP boost on the KITTI and COCO Traffic datasets, respectively.
\end{itemize}

\section{Related Works}
\subsection{Object Detection}
Object Detection is a fundamental task in computer vision and is more challenging than classification since it involves both localization and classification of objects in an image. Over the past decade, convolutional neural networks (CNNs) have achieved remarkable success in this domain. There are three main categories of CNN-based object detection methods: two-stage detectors, anchor-based one-stage detectors, and anchor-free one-stage detectors. Two-stage detectors, such as \cite{cai2018cascade,he2017mask,ren2015faster}, first generate region proposals using a region proposal network (RPN) and then classify and refine the bounding boxes in a second stage. Two-stage detectors tend to have higher accuracy compared to one-stage detectors at the expense of longer inference time. Anchor-based one-stage detectors \cite{lin2017focal,liu2016ssd,redmon2018yolov3} directly predict the category and bounding box of targets from feature maps and are thus more efficient than two-stage detectors. That being said, they use a large number of pre-defined anchor boxes as reference points, which results in additional computation. To reduce such computation, anchor-free one-stage detectors \cite{duan2019centernet,tian2019fcos,yang2019reppoints} directly predict the critical points and placements of objects without the use of anchor boxes, at the risk of sacrificing accuracy.


\begin{figure*}[h]
\begin{center}
\includegraphics[width=0.95\linewidth]{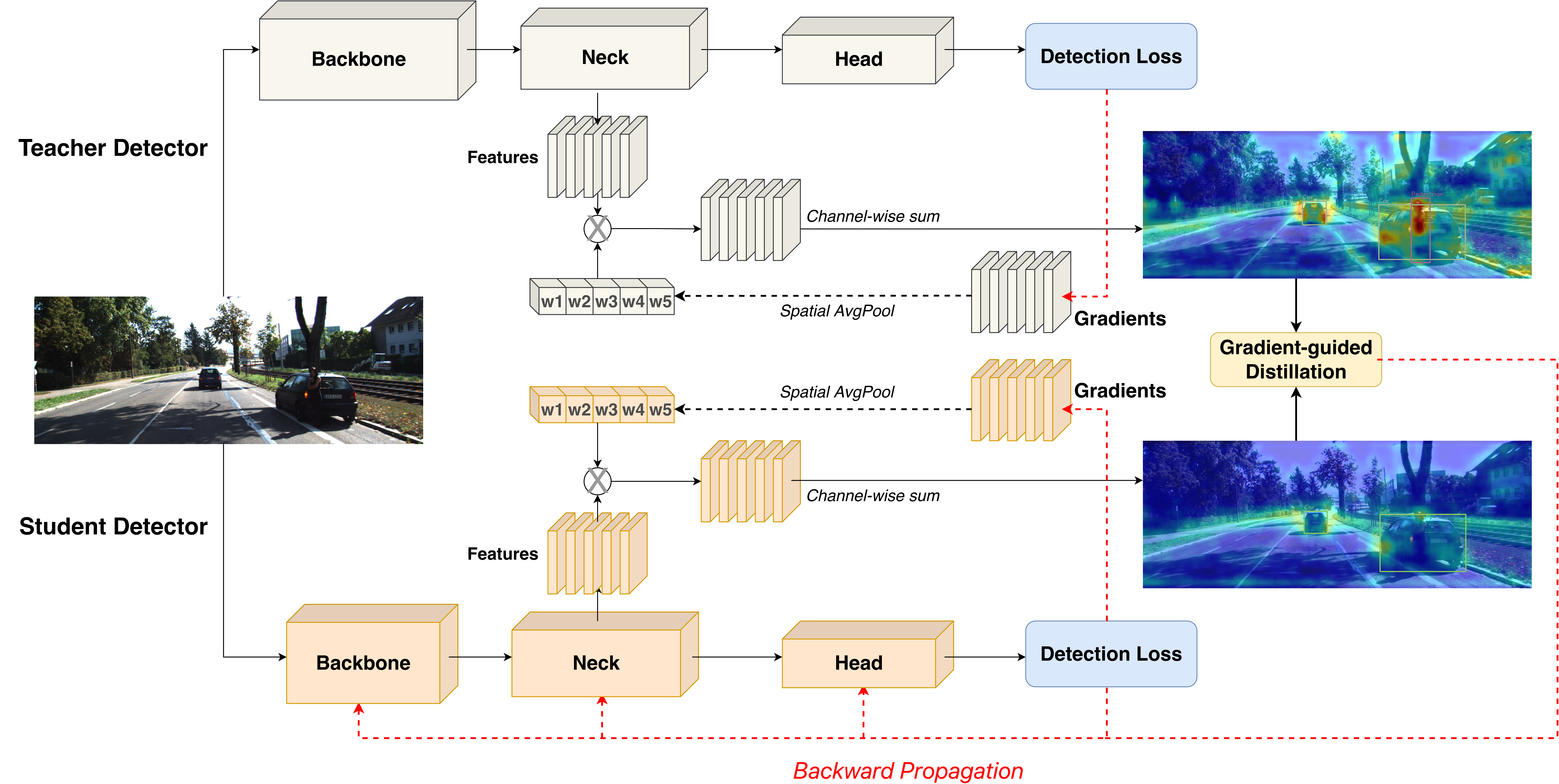}
\end{center}
\caption{
Illustration of the proposed Gradient-Guided Knowledge Distillation (GKD) method.
}
\label{fig:GKD}
\end{figure*}
\subsection{Knowledge Distillation}
Knowledge distillation is a model compression technique proposed by \cite{hinton2015distilling}. In its original version, the output probabilities or logits of a pre-trained teacher network serve as soft labels to guide the learning of a smaller student network for classification tasks. Since then, there have been many KD works (e.g., \cite{heo2019comprehensive,tung2019similarity,zagoruyko2016paying}) that further improve the vanilla KD's performance in classification tasks.
Relatively speaking, fewer works have applied knowledge distillation to object detection. \cite{chen2017learning} first apply knowledge distillation to object detection by distilling knowledge from the neck features, the classification head, and the regression head. Nevertheless, not all features in the teacher model are useful and relevant. Naively distilling all the features may mislead the student model. How to select the most valuable features for knowledge distillation in object detection is an active research area. \cite{li2017mimicking} choose the features sampled from the region proposal network (RPN) to improve the performance of the student model. \cite{wang2019distilling} propose the fine-grained mask to distill the regions near the ground-truth bounding boxes. \cite{sun2020distilling} utilize Gaussian masks to assign more importance to bounding boxes and surrounding regions for distillation. Such methods attempt to find the most informative spatial locations while ignoring the channel-wise feature selection. \cite{guo2021distilling} show that both the foreground and background play important roles for distillation, and distilling them separately benefits the student. \cite{dai2021general} distill the locations where the performances of the student and teacher differ most. All the above-mentioned methods try to infer the most informative spatial regions for knowledge distillation (e.g., the foreground or background). However, they do not consider the differences in importance across different feature channels and how the features contribute to the final detection. \cite{zhang2021improve} and \cite{yang2022focal} incorporate non-local modules and consider both spatial and channel attention. However, their feature importance is only based on the magnitude of activation, which is not directly related to final detection, either. 
In contrast to those works, we propose gradient-guided knowledge distillation, which assigns larger weights to features that contribute more to the final detection.


\section{Methodology}

Most state-of-the-art feature-based KD methods have the student model directly mimic the teacher model's plain features. Recently, some works like \cite{yang2022focal} and \cite{zhang2021improve} direct more focus to channels/locations that are highly activated.
Unlike previous approaches, we propose gradient-guided knowledge distillation (GKD) that gives special attention to knowledge contributing to the final detection performance. In addition, we will present how to incorporate bounding box and context information in multi-grained feature-based knowledge distillation.

\begin{figure*}[ht]
\begin{center}
\includegraphics[width=0.8\linewidth]{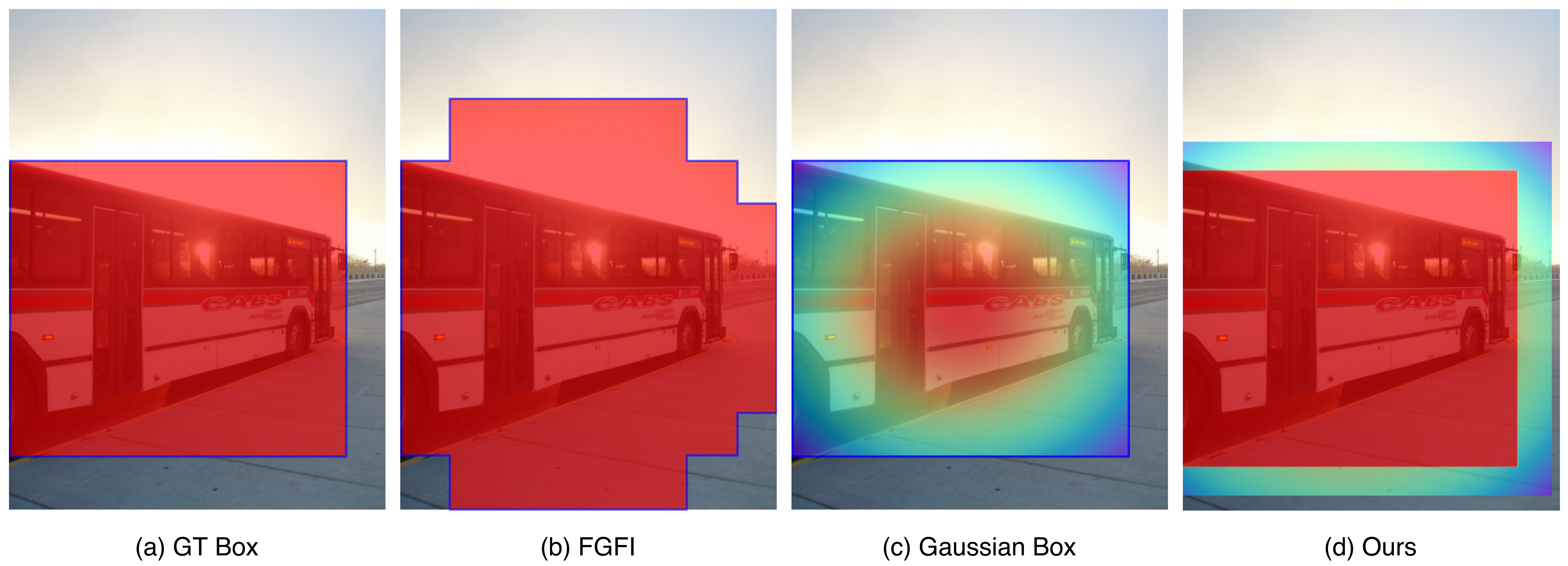}
\end{center}
\caption{
Popular attention regions for knowledge distillation in object detection. Different colors indicate different weights for different areas, with the red color representing the highest weights and the blue color representing the lowest. In contrast to other methods (a)-(c) \protect\cite{guo2021distilling,wang2019distilling,sun2020distilling}, our approach (d) focuses on foreground objects and their surrounding pixels with gradually diminishing weights.
}
\label{fig:Masks}
\end{figure*}
\begin{figure}[ht]
\begin{center}
\includegraphics[width=0.95\linewidth]{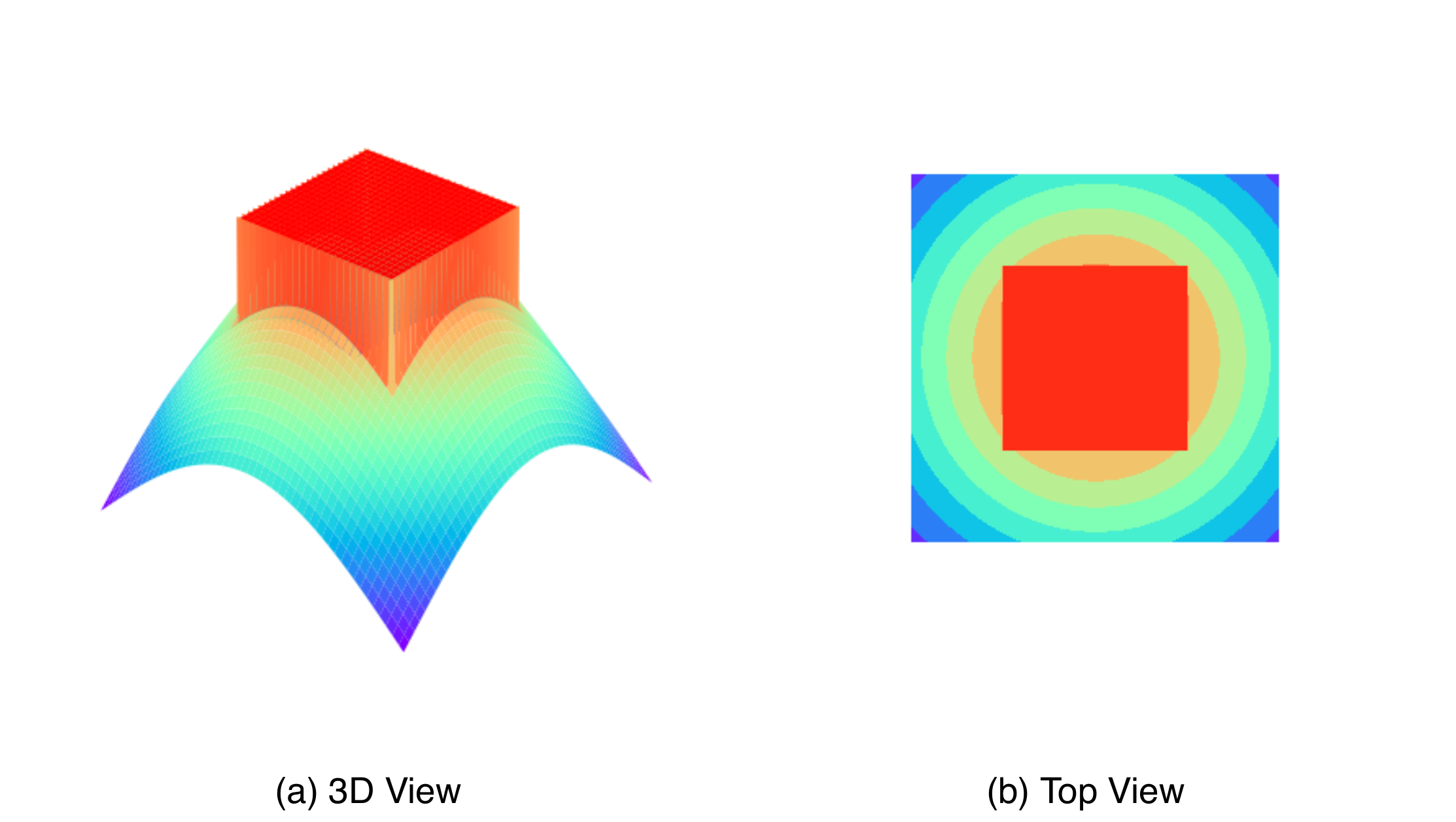}
\end{center}
\caption{
Views of our top-flattened Gaussian Mask. Different colors indicate different weights for different areas, with the red color representing the highest weights (ground truth area) and the blue color representing the lowest.
}
\label{fig:flattopgaussian}
\end{figure}

\subsection{Gradient-Guided Knowledge Distillation}
\label{GKD}




We propose to utilize the gradients of the detection loss with respect to features to represent the features' contribution to the final detection. The features corresponding to larger gradients are more influential on the decision making and thus they deserve more attention during the knowledge distillation process. Fig. \ref{fig:GKD} illustrates the general idea of our GKD and how it guides the student model to better learn the most valuable and relevant knowledge from the teacher.


Mathematically, we define the importance/weight of the $k$-th feature map in layer $l$ of a detector as:

\begin{equation}\label{eq:gradweight}
w_{k}^l = \frac{1}{WH}\sum_{i=1}^{W}\sum_{j=1}^{H}\frac{\partial \mathcal{L}_{task}}{\partial A_{i,j,k}^l}
\end{equation}

\noindent where $\mathcal{L}_{task}$ denotes the total detection loss (including bounding box regression loss and classification loss), $A_{i,j,k}$ is the single activation value at location $(i,j)$ in the $k$th feature map of the $l$th layer. We first calculate the gradients of $\mathcal{L}_{task}$, with respect to feature $A_{i,j,k}^l$. These gradients flowing back are global-average-pooled over the width and height dimensions (indexed by $i$ and $j$, with max value $W$ and $H$, respectively) to obtain the feature channel importance $w_{k}^l$. Then, we use $w_{k}^l$ to weigh the $k$th activation map $A_{k}^l$:

\begin{equation}\label{eq:weightedmap}
\widetilde{A_{k}^l} = w_{k}^l A_{k}^l
\end{equation}

\noindent where $\widetilde{A_{k}^l}$ is the $k$th gradient-weighted activation map of the $l$th layer. These maps are then linearly combined along the channel dimension (before ReLU and Norm) to obtain the final target map for distillation:

\begin{equation}\label{eq:targetmap}
M^l = Norm(ReLU(\sum_{k=1}^C \widetilde{A_{k}^l}))
\end{equation}

\noindent where $Norm$ represents the min-max normalization function and the $ReLU$ function removes negative values because we are only interested in the features that have a positive influence on the detection task. Negative pixels are likely those that belong to the background. 
Combining Eq. \ref{eq:gradweight}, \ref{eq:weightedmap}, and \ref{eq:targetmap}, we get:

\begin{equation}
    M^l = Norm(ReLU(\sum_{k=1}^{C}A^l_{k} \frac{1}{WH}\sum_{i=1}^{W}\sum_{j=1}^{H}\frac{\partial \mathcal{L}_{task}}{\partial A_{i,j,k}^l}))
\end{equation}

The same process can be applied to both the teacher model and the student model. The resulting target maps for the teacher and the student are $M^l_{\mathcal{T}}$ and $M^l_{\mathcal{S}}$, respectively. The goal of our gradient-guided knowledge distillation is to minimize the difference between the two target maps:

\begin{equation}
    \mathcal{L}_{GKD} = \frac{1}{HW}\sum_{l=1}^L\sum_{i=1}^W\sum_{j=1}^H\ |M^l_{i,j,\mathcal{T}} - M^l_{i,j,\mathcal{S}}|
\end{equation}

\noindent where $l$ indicates an intermediate layer. It ranges from 1 to $L$, with $L$ being the total number of intermediate layers of the student and teacher models being considered for distillation. We sum the absolute difference between $M^l_{i,j,\mathcal{T}}$ and $M^l_{i,j,\mathcal{S}}$. We use L1-norm loss instead of L2-norm loss because L2 can be more susceptible to outliers when there is a large discrepancy between the teacher and student models at the beginning of training. Using L1-norm loss encourages teacher-student consistency in more locations.

To handle objects of various scales, most state-of-the-art object detectors employ Feature Pyramid Networks (FPN) \cite{fpn} or its variant \cite{qiao2021detectors}. In our experiments, to enhance the transfer of knowledge across different scales, we choose the output layers of FPN as the target layers for distillation. 

\subsection{Bounding-box-aware Multi-grained Feature Imitation}

For KD of object detectors, the foreground and the background usually contain different amounts of useful information. The overwhelming amount of background information may mislead the knowledge distillation process. \cite{guo2021distilling} (Fig. \ref{fig:Masks}a) and \cite{yang2022focal} assign different weights to the foreground and the background. \cite{wang2019distilling} (Fig. \ref{fig:Masks}b) distills the anchor-covered regions around the foreground object. \cite{sun2020distilling} (Fig. \ref{fig:Masks}c) uses a Gaussian Mask to cover the ground truth bounding box for distillation. They either ignore the surrounding pixels or cover too many unnecessary regions. Unlike these approaches, we propose a top-flattened Gaussian mask $M_{i,j}$, which is defined as:

\begin{equation}\label{eq:flattopgaussian}
    M_{i,j}= \begin{cases}
    1,\quad & \text{if}\quad (i,j)\in o\\
    e^{-\frac{1}{2}\left(\frac{x-\bar x}{\bar x}+\frac{y-\bar y}{\bar y}\right)^2},\quad & \text{elif}\quad  (i,j)\in \hat o\\
    0,\quad & \text{otherwise}
\end{cases} 
\end{equation}
\vspace{0.1in}

\noindent where $o$ and $\hat o$ represent the regions inside ground truth bounding boxes and the regions surrounding them, respectively. $x$ and $y$ denote the width and height of the distillation region, which are set to be twice the width and height of the ground truth bounding box. (Fig. \ref{fig:flattopgaussian} show the 3D view and top view of our top-flattened Gaussian distribution). $(\bar x, \bar y)$ represents the center point of the ground truth bounding box. 
Eq.~\ref{eq:flattopgaussian} directs enough attention to the foreground while taking the surrounding pixels/regions into consideration as well. The surrounding pixels provide valuable contextual information for localizing the foreground object. Figure \ref{fig:Masks} illustrates the differences between our top-flattened Gaussian mask and previous methods.

To further improve the KD performance, we also incorporate spatial and channel attention (based on highly-activated features) \cite{yang2022focal} when distilling features. The spatial and channel attention masks can be defined as follows:

\begin{equation}\label{eq:spatialattention}
    M^S = W H\cdot softmax(\frac{\sum_{k-1}^C|A_k|}{C T})
\end{equation}

\begin{equation}\label{eq:channelattention}
    M^C = C \cdot softmax(\frac{\sum_{i=1}^W\sum_{j=1}^H|A_{i,j}|}{W H T})
\end{equation}
\vspace{0.05in}

\noindent where $A$ represents the plain feature, and $W,H,C$ are the width, height, and channel number of $A$ indexed by $i,j,k$, respectively. $T$ is the temperature hyper-parameter introduced by \cite{hinton2015distilling} to modulate the distribution. Based on Eq.~\ref{eq:flattopgaussian}, \ref{eq:spatialattention}, and \ref{eq:channelattention}, we propose our Bounding-box-aware Multi-grained Feature Imitation (BMFI) loss as follow:

\begin{align}\label{eq:bmfi}
    L_{BMFI} = & \sum_{k=1}^C\sum_{i=1}^H\sum_{j=1}^WM_{i,j}M^SM^C(A^{\mathcal{T}}_{i,j,k} - A^{\mathcal{S}}_{i,j,k})^2 \nonumber
    \\
    + & \alpha(|M^S_{\mathcal{T}} - M^S_{\mathcal{S}}| + |M^C_{\mathcal{T}} - M^C_{\mathcal{S}}|)
\end{align}
\vspace{0.015in}

\noindent where the subscript $\mathcal{T}, \mathcal{S}$ denotes the teacher and student detector, respectively. $\alpha$ is the hyper-parameters to balance the loss terms. 

By adding our Gradient-guided knowledge distillation loss from Sec.~\ref{GKD}, our total distillation loss is formulated as:

\begin{equation}
    L_{KD} =  L_{GKD} + \beta L_{BMFI}
\end{equation}

\noindent where $\beta$ is the hyperparameter that balances the contribution of two loss terms. $alpha$ and $beta$ are empirically set in our experiments to achieve the best validation results.


\section{Experiments and Results}
\subsection{Dataset}
\textbf{KITTI} \cite{Geiger2012CVPR} is a 2D-object detection dataset that includes seven different types of road objects. As suggested in \cite{KITTI_tran}, we group similar categories into one. Specifically, we perform the following modification to the original KITTI dataset:
\begin{itemize}
    \item Car $\leftarrow$ \textit{car, van, truck, tram}
    \item Pedestrian $\leftarrow$ \textit{pedestrian, person}
    \item Cyclist $\leftarrow$ \textit{cyclist}
\end{itemize}

It includes 7481 images with annotations. We split it into a training set and a validation set in the ratio of 8:2.

\textbf{COCO-Traffic} is a dataset containing 13 traffic-related categories. This dataset is obtained by selecting categories related to self-driving from MS COCO 2017 \cite{lin2014microsoft}. The COCO-Traffic dataset includes the following categories: 
\begin{itemize}
    \item \textbf{Road-related:} \textit{bicycle, car, motorcycle, bus, train, truck, traffic light, fire hydrant, stop sign, parking meter}
    \item \textbf{Others:} \textit{person, cat, dog}
\end{itemize}
We keep only images containing at least one road-related object to filter out those images that only contain indoor objects. The selection is applied to both the training and validation sets.



\subsection{Implementation Details}
All the detection experiments are conducted in the MMDetection framework \cite{mmdetection} using Pytorch \cite{Pytorch}. We employed Faster-RCNN \cite{ren2015faster} as a representative of two-stage detectors and chose Generalized Focal Loss (GFL) \cite{li2020generalized} as an example of one-stage detectors. The teacher and student models (without any knowledge distillation) were trained directly using the default configuration of MMDetection \cite{mmdetection}. The teacher models were based on a ResNet-101 backbone, and we tested two different student backbone architectures (i.e., ResNet-50 and ResNet-18). For comparison, we re-implemented the following state-of-the-art KD methods:

\begin{itemize}
    \item FGFI by \cite{wang2019distilling}, CVPR'19
    \item FKD by \cite{zhang2021improve}, ICLR'21
    \item GID by \cite{dai2021general}, CVPR'21
    \item DeFeat by \cite{guo2021distilling}, CVPR'21
    \item FGD by \cite{yang2022focal}, CVPR'22
\end{itemize}

All the competing knowledge distillation methods and our method are applied to FPN output layers. The temperature hyper-parameter $T$ is set to 0.5.
We adopt the inheriting strategy proposed in \cite{kang2021instance}, where the student model is initialized with the teacher's neck and head parameters. All the models are sufficiently trained to convergence with a SGD optimizer, an initial learning rate of 0.2, momentum of 0.9, and weight decay of 0.0001. All models are evaluated in terms of mean averaged precision (mAP) with 0.5 as the Intersection over Union (IoU) threshold.

\subsection{Experiment Results}
\begin{table*}[tbh!]
\footnotesize
\centering
\begin{tabular}{c|M{1.9cm}M{1.9cm}|M{1.9cm}M{1.9cm}}
\specialrule{0.05em}{0pt}{1pt}
\multirow{2}{*}{\backslashbox{KD methods}{Student backbones}} &\multicolumn{2}{c|}{ResNet-50}    &\multicolumn{2}{c}{ResNet-18} \\\cline {2-5}
 & \rule{0pt}{2.2ex} KITTI          & \rule{0pt}{2.2ex} COCO Traffic   & \rule{0pt}{2.2ex} KITTI           & \rule{0pt}{2.2ex} COCO Traffic    \\\hline
Teacher (w ResNet-101)                           & 89.4  & 71.8  & 89.4  & 71.8 \\
Student-baseline                                 & 85.1  & 67.7  & 81.9  & 61.9 \\ \hline   
FKD \cite{zhang2021improve}                      & 86.4  & 69.5  & 84.4  & 62.6 \\
GID \cite{dai2021general}                        & 86.1  & 69.3  & 84.6  & 63.7 \\
DeFeat \cite{guo2021distilling}                  & 85.4  & 69.3  & 83.3  & 62.7 \\
FGD \cite{yang2022focal}                         & 89.2  & 71.0  & 86.7  & 65.9 \\
FGFI \cite{wang2019distilling}                   & 84.4  & 68.6  & 82.6  & 62.4 \\
\rowcolor{lightgray!40}
Our GKD                                              & 90.0  & 69.5  & 88.1  & 66.2 \\
\rowcolor{lightgray!40}
Our GKD-BMFI                                         & \textbf{90.3}  & \textbf{71.2}  & \textbf{88.7}  & \textbf{66.6} \\
\hline
\end{tabular}
\caption{Performance (mAP) of different distillation methods with GFL detector \protect\cite{li2020generalized} on the KITTI and COCO traffic datasets. (The teacher model and the student-baseline are non-distillation GFL models with ResNet-101 and ResNet-50\slash18 as backbones, respectively.) The highest mAP in each column is highlighted.}
\label{tab:GFL}
\end{table*}
\begin{table*}[h!]
\footnotesize
\centering
\begin{tabular}{c|M{1.9cm}M{1.9cm}|M{1.9cm}M{1.9cm}}
\specialrule{0.05em}{0pt}{2pt}
\multirow{2}{*}{\backslashbox{KD methods}{Student backbones}}  &\multicolumn{2}{c|}{ResNet-50}    &\multicolumn{2}{c}{ResNet-18} \\
\cline {2-5}
 & \rule{0pt}{2.2ex} KITTI          & \rule{0pt}{2.2ex} COCO Traffic   & \rule{0pt}{2.2ex} KITTI           & \rule{0pt}{2.2ex} COCO Traffic    \\\hline
Teacher (w ResNet-101)                          & 89.3  & 67.9  & 89.3  & 67.9 \\
Student-baseline                                & 88.9  & 67.5  & 84.1  & 63.1 \\ \hline
FKD \cite{zhang2021improve}                     & 89.0  & 67.8  & 87.2  & 65.3 \\ 
FGD \cite{yang2022focal}                        & 88.9  & 67.7  & 87.0  & 64.1 \\
\rowcolor{lightgray!40}
Our GKD                                             & 90.6  & 69.8  & 89.0  & 66.5 \\
\rowcolor{lightgray!40}
Our GKD-BMFI                                        & \textbf{90.8}  & \textbf{70.3}  & \textbf{89.1}  & \textbf{66.9} \\
\hline
\end{tabular}
\caption{Performance (mAP) of different distillation methods with Faster R-CNN detector \protect\cite{ren2015faster} on the KITTI and COCO traffic datasets. (The teacher model and the student-baseline are non-distillation Faster-RCNN models with ResNet-101 and ResNet-50/18 backbones, respectively.) The highest mAP in each column is highlighted.}
\label{tab:Faster-rcnn}
\end{table*}

\begin{table}[t]
    \centering
    \begin{tabular}{c|c|ccccc}
    \toprule
    Detector & \multicolumn{6}{c}{GFL-ResNet-50}\\
    \midrule
    \multirow{3}{*}{Modules} & GKD & $\times$ &$\times$ &$\times$ & $\checkmark$ & $\checkmark$\\
                             & MASK& $\times$ & $\times$ & $\checkmark$ & $\checkmark$ & $\times$\\
                             & MFI & $\times$ & $\checkmark$ & $\checkmark$ & $\checkmark$ & $\times$ \\\midrule
    Results & mAP & 85.1 & 88.5 & 89.7 & \textbf{90.3} & 90.0\\
    \bottomrule
    \end{tabular}
    \caption{Ablation study of the three different components of our GKD-BMFI. This ablation study is conducted on the KITTI dataset using GFL with a ResNet-50 backbone. GKD: gradient-guided knowledge distillation (with no bells and whistles), MASK: bounding-box-aware Gaussian mask, MFI: multi-grained feature imitation ($M_S$ and $M_C$ related).}
    \label{tab:ablation_study}
\end{table}

In our experiments, we evaluated the performance of our proposed gradient-guided knowledge distillation (GKD) method against several state-of-the-art knowledge distillation methods on the KITTI and COCO-Traffic datasets using both single-stage (e.g., GFL) and two-stage (e.g., Faster RCNN) object detectors. The results on the single-stage and two-stage detectors are shown in Table \ref{tab:GFL} and Table \ref{tab:Faster-rcnn}, respectively.

According to Table \ref{tab:GFL}, we can see that our GKD method provides a significant boost in mAP for single-stage student detectors. Specifically, when using a ResNet-50 backbone, our GKD method achieved 4.9 and 2.3 mAP improvement on the KITTI and COCO-Traffic datasets, respectively. Similarly, when using a ResNet-18 backbone, our GKD method achieved 6.2 and 4.3 mAP improvement on the KITTI and COCO-Traffic datasets, respectively. Our GKD-BMFI, which incorporates Bounding-box-aware Multi-grained Feature Imitation, outperforms all student baseline models and other state-of-the-art distillation methods. For example, on the KITTI dataset, our GKD-BMFI outperforms FGD \cite{yang2022focal} by 1.1 mAP with a ResNet-50 backbone and 2.3 mAP with a ResNet-18 backbone. On the COCO-Traffic dataset, it surpasses other five different KD methods by an average of 4.4 mAP with a ResNet-50 backbone and 1.7 mAP with a ResNet-18 backbone.

As shown in Table \ref{tab:Faster-rcnn}, our proposed GKD method is also effective for two-stage detectors. Specifically, when utilizing a ResNet-50 backbone on the COCO-Traffic dataset, our GKD method demonstrates a remarkable improvement of 2 mAP over the student-baseline and outperforms other state-of-the-art distillation methods, including \cite{zhang2021improve} and \cite{yang2022focal}, by an average of 1.75 mAP. In addition, our GKD-BMFI can further improve the distillation performance. For example, when comparing to the student-baseline with a ResNet-18 backbone on the KITTI dataset, our GKD-BMFI method demonstrates an impressive improvement of 5 mAP.

\subsection{Qualitative Analysis}

\begin{figure*}[btp]
\begin{center}
\includegraphics[width=0.95\linewidth]{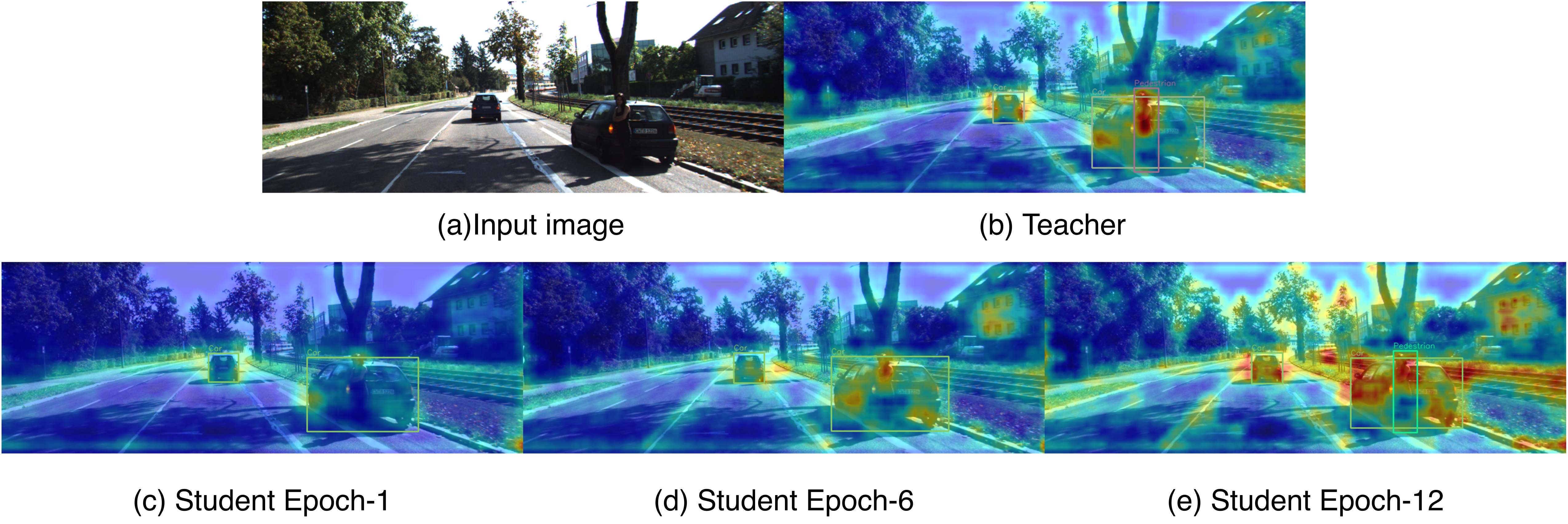}
\end{center}
\caption{
Visualization of the gradient-guided masks from the teacher detector and different training stages of the student detector using GKD. This experiment is conducted on the KITTI dataset using the GFL detector. Different colors indicate different attention levels, with the red color representing the highest attention and the blue color representing the lowest.
}
\label{fig:GKD_VIS}
\end{figure*}
\begin{table*}[tbh]
\footnotesize

\centering
\begin{tabular}{cccc}
\toprule
\multicolumn{1}{c}{Model} & \multicolumn{1}{c}{Backbones} & Parames(M)  & GFLOPs \\
\midrule
\multirow{2}{*}{GFL \cite{li2020generalized}}  & ResNet-101 & 51.03 & 13.79   \\ 
& ResNet-50 & 32.04 & 10.05 \\ 
& ResNet-18 & 19.09 & 7.61 \\ \midrule
\multirow{2}{*}{Faster R-CNN \cite{ren2015faster}}  & ResNet-101 & 60.13 & 27.09 \\
& ResNet-50 & 41.13 & 23.36\\
& ResNet-18 & 28.13 & 20.77\\
\bottomrule
\end{tabular}
\caption{Model complexity (with 224$\times$224 input resolution)}
\label{tab:efficiency}
\end{table*}
In this subsection, we visualize the gradient-guided masks from the teacher detector and different stages of the student detector, as shown in Fig. \ref{fig:GKD_VIS}. This example from our experiments on the KITTI dataset using the GFL detector. By comparing the gradient-guided masks between the teacher and the students at different training stages, we can observe the student's gradual learning process and see how it tries to follow the teacher's guidance. According to the figure, the teacher detector (Fig. \ref{fig:GKD_VIS} (b)) focuses on the objects in the image (e.g cars and pedestrians) more accurately than the student detector that has only been trained for one epoch (Fig. \ref{fig:GKD_VIS} (c)). However, as our gradient-guided knowledge distillation process goes on, we can see that the student's attention gradually becomes more similar to the teacher's, as seen in Fig. \ref{fig:GKD_VIS} (d). In Fig. \ref{fig:GKD_VIS} (e), we can see that the student even develops some new high-attention areas (e.g., the smaller-scale car in front of the vehicle). This potentially explains why our much smaller distilled model even surpasses the teacher model in this case (90.3 mAP vs. 89.4 mAP).

\subsection{Ablation Study}
As we see from previous subsections, our GKD-based methods can improve the performance of the student baseline by large margins. To analyze which components of our method contributes most to the mAP boost, we perform an ablation study in this subsection. We perform the ablation study on the KITTI dataset using the GFL \cite{li2020generalized} detector with a ResNet-50 backbone. We consider the following three components in this study: our Gradient-guided Knowledge Distillation (GKD, without bells and whistles), bounding-box-aware mask (MASK), and Multi-grained Feature Imitation (MFI) methods.
The results are shown in Table \ref{tab:ablation_study}. According to the table, all three components play a positive role in the mAP boost, but the GKD with no bells and whistles makes the most contribution. To be more specific, GKD alone can improve the baseline mAP from 85.1 to 90.0. The combination of the three components results in the highest mAP of 90.3 (a 5.2 mAP improvement).
From Table \ref{tab:ablation_study}, we can also see that using only the MFI component results in a 3.4 mAP improvement. By incorporating the bounding-box-aware mask (MASK) into MFI, we get BMFI (as described in Eq. \ref{eq:bmfi}), which results in a 4.6 mAP improvement over the student detector.

\subsection{Complexity}

In addition to mAP performance, we also compared different architectures' efficiency in terms of FLOPs\footnote{we count one multiplication and one addition operation as one FLOP.} and the number of parameters. The results are shown in Table \ref{tab:efficiency}. According to the table, our distilled models with the smaller backbones (ResNet-50 or ResNet-18) are more efficient than the corresponding teacher models with larger ResNet-101 backbones. 
In addition to the previously mentioned promising mAPs, our ResNet-50/18 distillation model enjoys an average of 34.41\%/67.90\% reduction in model size (number of parameters) and an average of 20.22\%/33.70\% savings in FLOPs.

\section{Conclusion}

In this paper, we have proposed a novel gradient-guided knowledge distillation (GKD) method. It leverages the gradients of the detection loss w.r.t. feature maps to identify valuable and relevant knowledge for knowledge distillation. Our GKD gives special attention to feature maps contributing more to the final detection. In addition, we have presented bounding-box-aware multi-grained feature imitation (BMFI) to further improve the distilled model's performance. Experiments on the KITTI and COCO-Traffic datasets, using various detectors and backbones, demonstrate our method's efficacy. The qualitative analysis shows that our gradient-guided knowledge distillation allows the student to get similar or even more informative attention maps than the teacher.

\bibliographystyle{named}
\bibliography{ijcai23}

\end{document}